\documentclass[10pt,twocolumn,letterpaper]{article}

\usepackage{cvpr}
\usepackage{times}
\usepackage{epsfig}
\usepackage{graphicx}
\usepackage{amsmath}
\usepackage{amssymb}
\usepackage{multirow}
\usepackage{cellspace}
\usepackage{tabularx}
\usepackage{makecell}
\usepackage{xcolor}
\usepackage{float}
\usepackage{mathptmx}

\usepackage{xspace}
\usepackage{color}
\usepackage{colortbl}

\usepackage{booktabs}                   

\definecolor{gray}{rgb}{0.35,0.35,0.35}
\definecolor{yellow}{rgb}{1,1,0.25}
\definecolor{MyBlue}{rgb}{0,0.2,0.8}
\definecolor{MyRed}{rgb}{0.8,0.2,0}
\definecolor{Red}{rgb}{1,0,0}
\definecolor{LightCyan}{rgb}{0.88,1,1}
\definecolor{MyGreen}{rgb}{0.0,0.5,0.1}
\definecolor{LightGray}{rgb}{0.9,0.9,0.9}

\def\red#1{\textcolor{red}{#1}}
\def\blue#1{\textcolor{blue}{#1}}

\newlength\paramargin
\newlength\figmargin
\newlength\secmargin

\usepackage{graphicx}
\graphicspath{{figures/}}

\usepackage{array}
\newcolumntype{L}[1]{>{\raggedright\let\newline\\\arraybackslash\hspace{0pt}}m{#1}}
\newcolumntype{C}[1]{>{\centering\let\newline\\\arraybackslash\hspace{0pt}}m{#1}}
\newcolumntype{R}[1]{>{\raggedleft\let\newline\\\arraybackslash\hspace{0pt}}m{#1}}

\long\def\ignorethis#1{}
\def\@onedot{\ifx\@let@token.\else.\null\fi\xspace}

\renewcommand{\paragraph}[1]{\noindent\textbf{#1}}
\usepackage{adjustbox}
\usepackage{cite}
\usepackage[breaklinks=true,bookmarks=false]{hyperref}

\cvprfinalcopy 

\def\cvprPaperID{****} 

\begin{document}

\title{SRPGAN: Perceptual  Generative Adversarial Network for Single Image \\
Super Resolution}

\author{Bingzhe Wu\quad Haodong Duan\quad Zhichao Liu\quad Guangyu Sun\\
Peking Univeristy\\
Beijing, 100871, China\\
{\tt\small \{wubingzhe,duanhaodong,liuzceecs,gsun\}@pku.edu.cn}
}

\maketitle

   \begin{abstract}
      Single image super resolution (SISR) is to reconstruct a high resolution image from a single low resolution image. The SISR task has been a very attractive research topic over the last two decades.
In recent years, convolutional neural network (CNN) based models have achieved great performance on SISR task. Despite the breakthroughs achieved by using CNN models, there are still some problems remaining unsolved, such as how to recover high frequency details of high resolution images. Previous CNN based models always use a pixel wise loss, such as l2 loss. Although the high resolution images constructed by these models have high peak signal-to-noise ratio (PSNR), they often tend to be blurry and lack high-frequency details, especially at a large scaling factor. In this paper, we build a super resolution perceptual generative adversarial network (SRPGAN) framework for SISR tasks. In the framework, we propose a robust perceptual loss based on the discriminator of the built SRPGAN model.
We use the Charbonnier loss function to build the content loss and combine it with the proposed perceptual loss and the adversarial loss. Compared with other state-of-the-art methods, our method has demonstrated great ability to construct images with sharp edges and rich details. We also evaluate our method on different benchmarks and compare it with previous CNN based methods. The results show that our method can achieve much higher structural similarity index (SSIM) scores on most of the benchmarks than the previous state-of-art methods.
   \end{abstract}

   \section{Introduction}
   \begin{figure*}[h]
\centering
\begin{tabular}{cc}
\textbf{Generator}& \textbf{Discriminator}\\
\vspace{+0.2cm}
\includegraphics[width = 8cm]{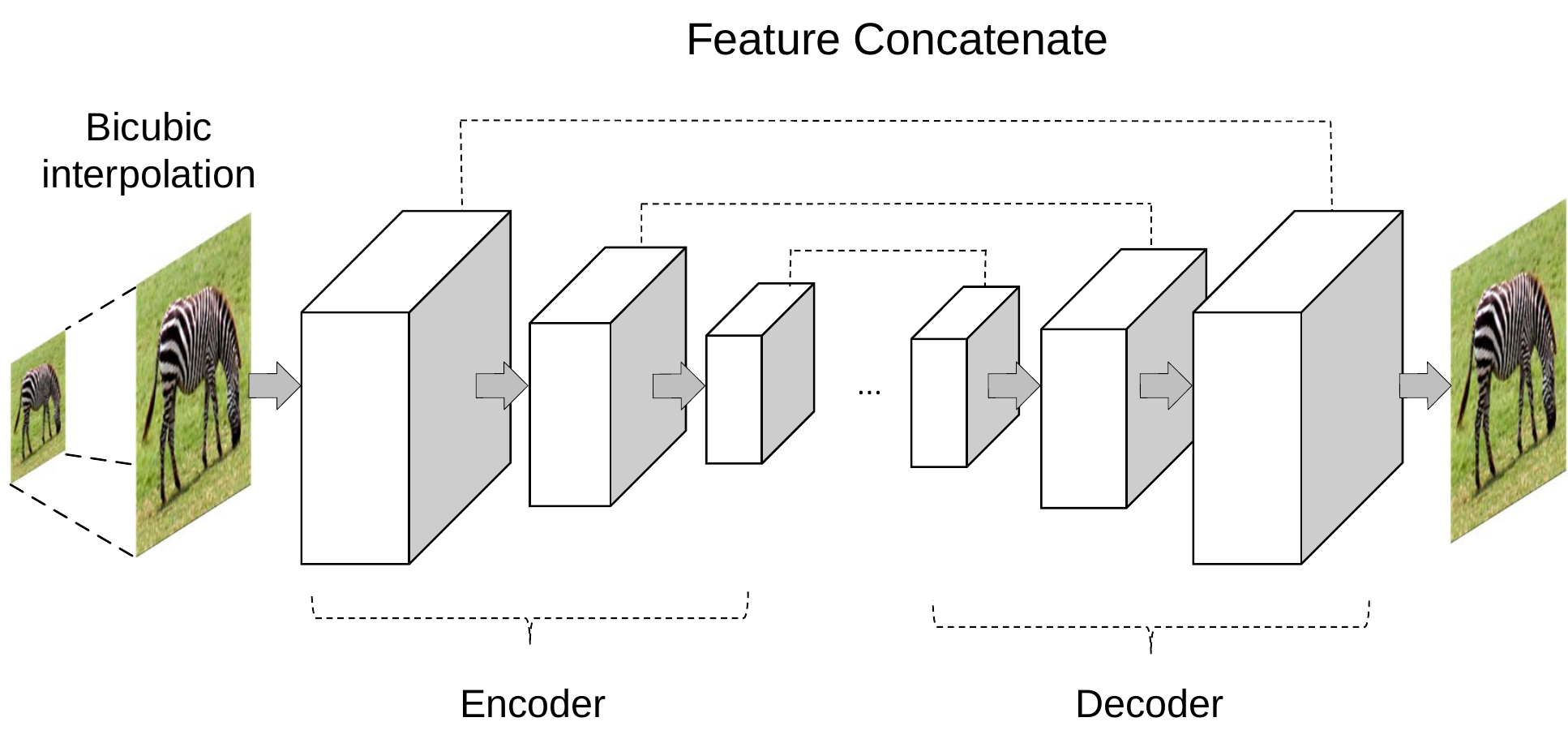}&
\includegraphics[width = 8cm, height = 4cm]{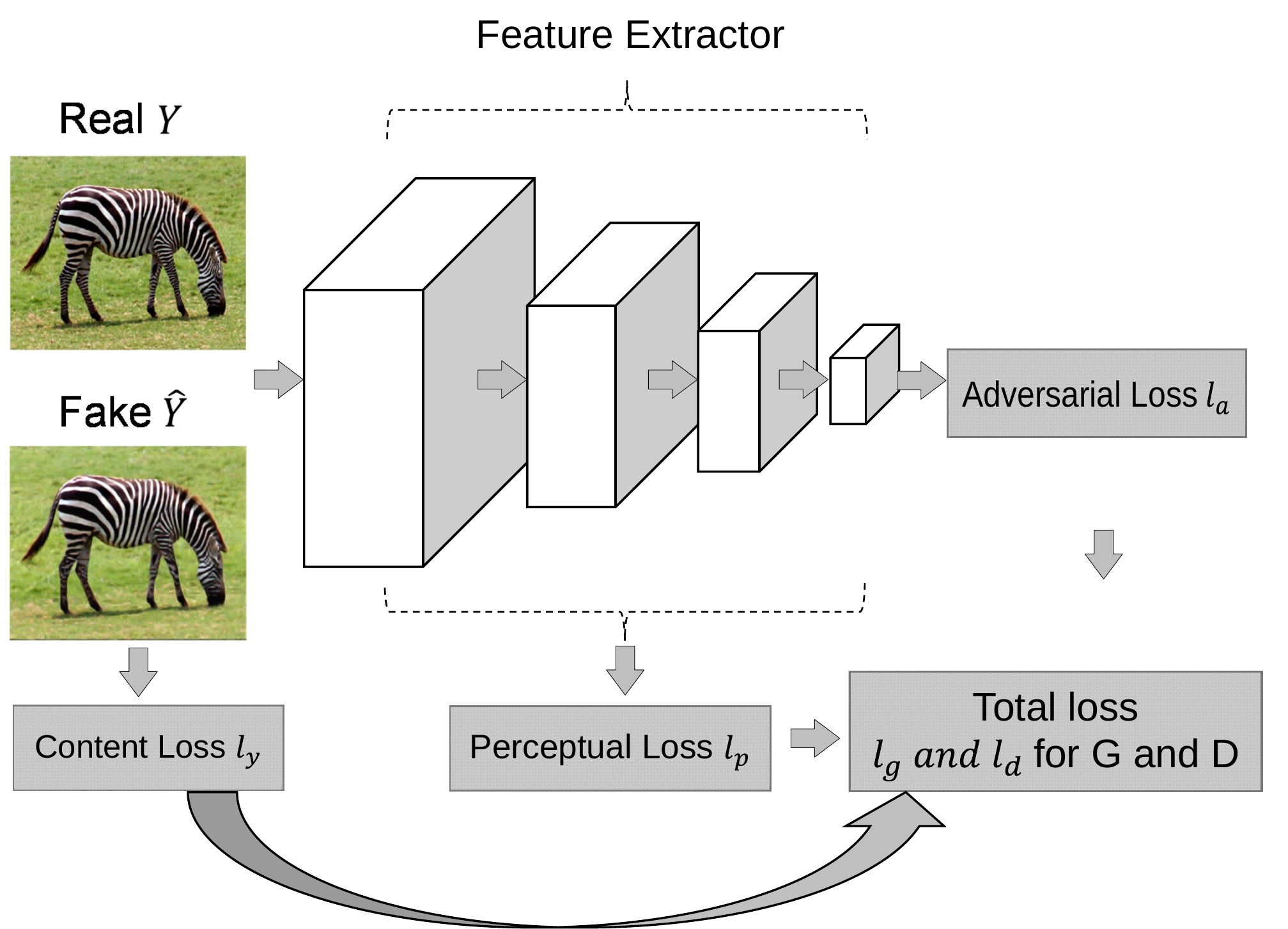}\\
\end{tabular}
\caption{The structure of the SISR framework(SRPGAN). Our Framework consists of a generator $G$ (left) and a discriminator $D$ (right). The generator part generates
high resolution image from low resolution one. The discriminator takes the generated image and the ground truth image as inputs to extract features of these inputs.
Based on these extracted features, the discriminator can build the objective functions for $D$ and $G$.}
\label{framewroks}
\end{figure*}

Single image super resolution (SISR) is a well defined problem in computer vision area. It tries to reconstruct a high resolution image from a single low resolution image. It has
been a very attractive research topic over the last two decades \cite{gerchberg1974super} \cite{aghajan1993sensor} \cite{glasner2009super}. Since SISR can restore some high frequency
details, it has been applied to many practical applications such as medical imaging \cite{shi2013cardiac}, satellite imaging \cite{thornton2006sub}, and face
identification \cite{bilgazyev2011improved}, where rich details are greatly desired.

In recent years, CNNs have shown powerful ability to learn highly non-linear transformations. Due to their powerful
learning ability, the CNN based methods are widely used for SISR tasks and have achieved remarkable progress. Despite those breakthroughs brought by the CNN based methods, there are still some critical problems, which remain largely unsolved, such as how to recover high resolution image with high perceptual quality and high frequency details. A common objective function of previous CNN based methods is the pixel wise loss function between the reconstructed and the ground truth high resolution images. The most commonly used
pixel wise loss is $l2$ loss. A method based on $l2$ loss is to minimize the mean square loss and maximize the peak signal-to-noise ratio (PSNR), which is a common measure used to evaluate SISR algorithms. Although such a method leads to high performance on PSNR metric, the images always tend to be blurry and over-smoothing \cite{Sonderby:2016ta}  \cite{dahl2017pixel}. Some recent literatures have pointed out that the pixel wise loss based methods failed to build multimodal distribution \cite{Sonderby:2016ta} \cite{dahl2017pixel} \cite{yang2010image}. There are two different approaches to solve this issue. One approach is to use a different constructed method, for example PixelCNN\cite{dahl2017pixel},~to build dependencies between different pixels. The other is using perceptual loss and adversarial learning to generate more realistic images \cite{Ledig:2016vc}. In this paper, we focus on the latter approach. 

In this work, we propose a general SISR framework (SRPGAN), which is based on the Image-to-Image
model \cite{isola2016image}. The start point of the proposed framework is the generative adversarial network (GAN). Unlike some previous methods \cite{Ledig:2016vc}, which uses a classification network to generate the perceptual loss, we use the features obtained by the discriminator network to build a more robust perceptual loss. We further design the adversarial loss and the content loss to build the final objective function. We also propose to use the Charbonnier loss function as the content loss function, which is different from the previous methods. In respect of the network architecture, we propose to replace the batch normalization layer with the instance normalization layer. We evaluate our method on most used benchmarks with a large upscaling factor. Our method outperforms other previous methods with SSIM score on most of benchmarks. Beside the quantitative evaluation, our method has also demonstrated great ability to reconstruct images with rich details and high perceptual quality.

The rest of this paper is organized as follows. The related work part summarizes the previous related works briefly. The methods section describes the framework details and the proposed individual loss functions. The quantitative evaluation and results visualization can be found in the experiments section.

   \section{Related Works}
    From the great performance achieved by the deep convolutional neural network at ImageNet challenge, various CNN based methods are applied to the super-resolution problem.
SRCNN is the first paper that applies CNN to the single image super-resolution problem \cite{Dong:2014td}. SRCNN is a simple model with three convolutional layers working for feature extraction, non-linear mapping, and image reconstruction. This method learns the end-to-end transformation between low and high resolution images. 

Based on the results of SRCNN, the authors accelerated the previous model by using an hour-glass shape CNN structure, and achieved a better SR performance and a real-time SR model with the rate higher than 24 fps on a generic CPU \cite{Dong:2016tw}. To achieve a real-time model for SR, they replaced the bicubic interpolation part in the previous model with deconvolution layers. After removing bicubic interpolation, this model can learn directly from the low resolution image. FSRCNN model only includes convolution layers and deconvolution layers. The convolution layers share the weights for different upscaling factors. With weight sharing, FSRCNN is able to deal with various scales using a single model.

Because of its success at the ImageNet challenge, the deep architecture similar with VGG net was proposed for large receptive fields in the VDSR (Very Deep network for Super-Resolution)~\cite{Kim:2015wv}. The convergence rate is the main limitation of the deeper model. The VDSR tries to use a higher learning rate of $10^{-1}$ while SRCNN used a learning rate of $10^{-5}$. Due to the high learning rate, it can be easier to diverge. By using the gradient clipping, the VDSR can be controlled strictly. The VDSR included the concept of residual learning to generate final output results. Due to the property of SISR problem, the output image is quite similar to the input image. With the concept of residual learning, the input image is added to the output of the model before making the final output image. As a result, the model can focus on the detail with high-frequency components. 
Besides, DRCN(Deeply-Recursive Convolutional Network) based on the VDSR model added recursion connections for weight sharing and model compression with only 5 layers \cite{Kim:2015wa}. The LapSRN is the one of the most recent frameworks for SISR problem \cite{lai2017deep}. The model includes a cascaded framework of feature extraction and image reconstruction parts using laplacian pyramids. And the Charbonnier loss function is used instead of the $l2$ loss function.

By replacing deconvolution layers with sub-pixel convolutional layers, the total computational complexity can be reduced dramatically. The new operator can also generate a cleaner image without checkerboard artifacts. With the efficient operation, ESPCN(Efficient Sub-Pixel Convolutional Neural Network) has achieved significant x10 speed up which can be applied for SR operation of HD videos  on a single GPU \cite{Shi:2016tm}.

The methods we mentioned above are always based on pixel wise loss, such as $l2$ loss. Although such a method leads to a high PSNR score, the images constructed by that always tend to be blurry and over-smoothing. Some recent works also have point out that pixel wise loss fails to capture  multimodal distribution \cite{dahl2017pixel}. There are two approaches to solve this issue. One approach is to use different network
structures to construct high resolution images. For example,
in \cite{dahl2017pixel}, the authors proposed
the PixelCNN to capture the dependencies between different pixels.
Another approach is to combine the perceptual loss and GAN model to generate more realistic and sharper image. In this paper, we mainly focus on the latter approach.
There are also some recent papers which focus on the perceptual loss. In \cite{Johnson:2016wm}, the authors firstly introduced the perceptual loss based on
VGG classification network for the style transformation and super resolution. The SRGAN method combines perceptual loss and adversarial loss for photo-realistic image \cite{Ledig:2016vc}. They address that the pixel wise loss does not capture the perceptual difference between ground truth images and output images. However, 
the perceptual loss in these paper is based on the VGG classification network, such 
a naive classification network cannot capture the desired high frequency details in super resolution tasks and will introduce extra computation. To this end, the VGG perceptual loss is not a suitable metric for SISR.
In this paper, we try to build a more robust perceptual loss to get
higher perceptual quality.

    \section{Methods}
    \subsection{SISR Framework}
We build our single image super resolution framework on Image-to-Image model\cite{isola2016image}. Our framework consists of an image generator $G$ and a discriminator $D$. The generator is trying to transform the image  in the domain generated by bicubic upsampling to the image in the ground truth high resolution image domain. 
The discriminator is trying to extract the features of the input high resolution images and the constructed images. Based on the features obtained by the discriminator, we can get the adversarial loss and the perceptual loss. 
Finally, we combine these two loss functions with the content loss to build the total objective functions of $D$ and $G$. The overall framework is illustrated in Figure\ref{framewroks}. 

Considering a single low-resolution image, we firstly upscale it by the specified factor using bicubic interpolation for further computing. Then, the generator network takes the interpolated image as the input and maps it to a high resolution image. Our final goal is to train a generator network $G$ that can generate high resolution image that is as similar as possible to the ground truth high resolution image. To achieve this, we construct a robust loss using the output and the intermediate features obtained by the discriminator $D$. Additionally, we design a content loss which can be used for evaluating the similarity between the generated image and the ground truth image. The individual loss functions are described with more details in the following subsection.
\subsubsection{Network architecture}
Our start point is the Image-to-Image model\cite{isola2016image}, we further tailor the Image-to-Image model for the SISR task. To the best of our knowledge, this is the
first paper that attempts to apply the image-to-image model to the SISR task.

\begin{figure}[!h]
\centering
\includegraphics[width = 4cm]{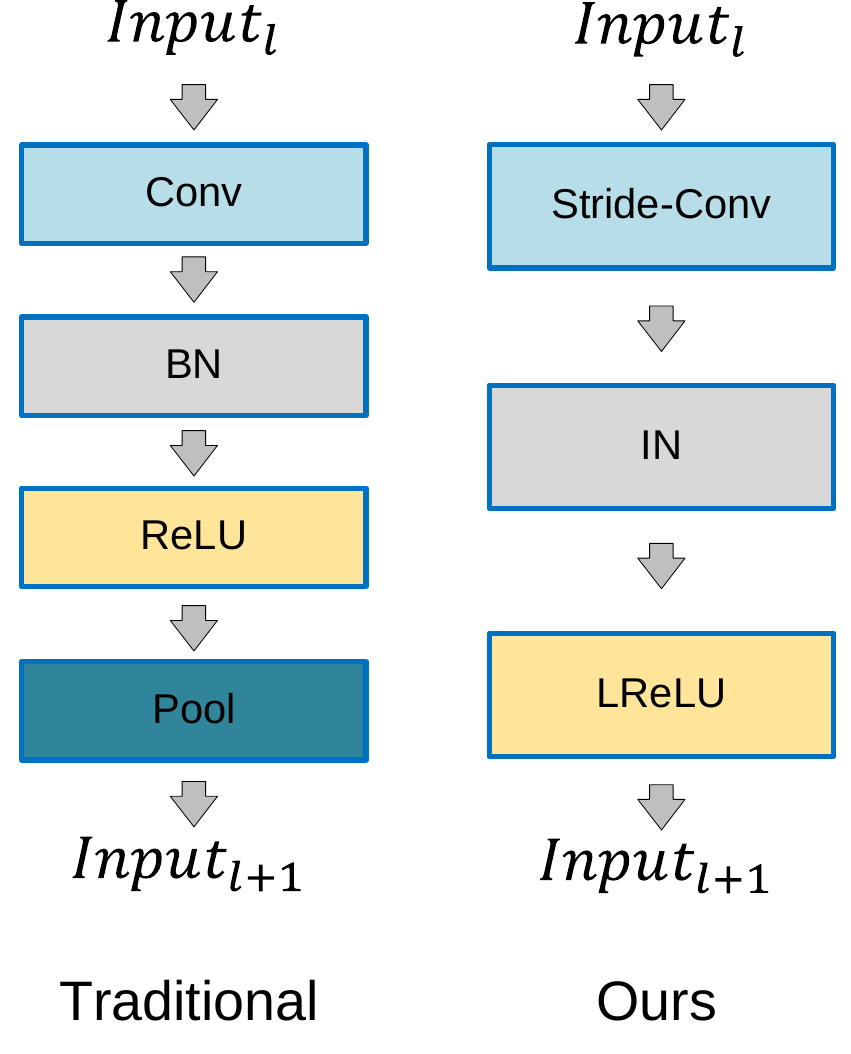}
\caption{Comparison of a conv block in a traditional GAN model and that in our model. Our model replaces the pooling layer with a stride-convolution layer. And we use the instance normalization (i.e. IN), not the batch normalization.}
\label{conv_layer}
\end{figure}

The generator $G$ is the core of the whole framework. The structure of $G$ illustrated in Figure\ref{framewroks}(left) has an encoder-decoder shape. We add skip connections following the general shape
of the U-Net \cite{ronneberger2015u} to combine the local and global information. Specifically, in our generator, skip connection is implemented by concatenating features obtained by layer $i$ and layer $n-i$,
where n is the total number of the convolution layers. The encoder part of $G$ consists of a stack of convolution layers. More Specifically, we use convolution layers with small
$3\times 3$ kernels. Following the previous work \cite{Ledig:2016vc}, we use the stride convolution to reduce the image resolution in each encoder layer, instead of the max pooling. 
Further more, we replace the batch normalization layers with the instance normalization layers\cite{ulyanov2016instance} to achieve better performance. In Figure\ref{conv_layer}, we 
show the difference between the convolution blocks in a traditional GAN model and those in our model.
We increase the resolution of the input features with transpose convolution layers in the decoder part. For the activation functions, we use the LeakyReLU activation functions in all encoder and decoder layers. 
We build the patch discriminator network following \cite{isola2016image}. Compared with the traditional discriminator, the patch discriminator tries to classify whether each patch in an image is real or fake instead of the whole image. Such a discriminator can restrict the GAN model to focus on the high frequency details. And the existence of the content loss can make sure the correctness of the low frequency part. The detail of the loss functions can be found in the following subsection. In the convolution blocks of the discriminator, we remove the batch normalization layers directly.
\begin{figure*}[!ht]
\centering
\includegraphics[width = 16cm]{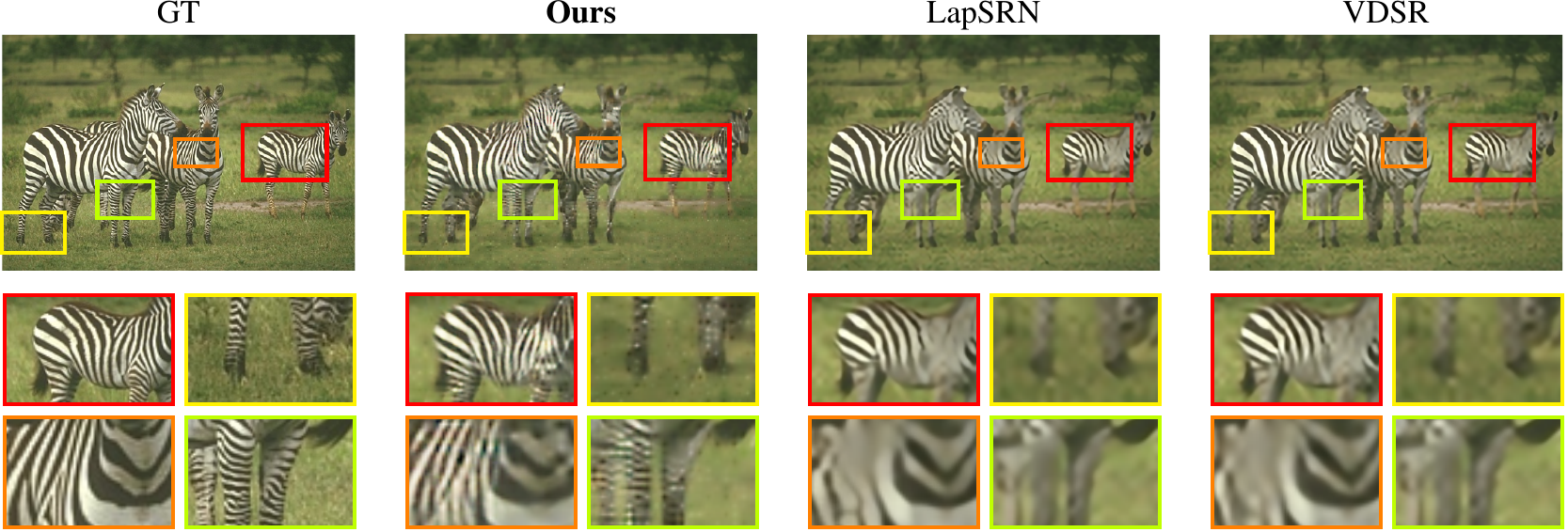}
\caption{Reconstruction results [$4\times$ upscaling] of \textbf{Ours} and recent state-of-the-art methods(LapSRN and VDSR). We use color boxes to highlight sub regions which contain rich details. We magnify the sub regions in the bellow boxes to show more details. From the sub region images, we can see that our method has stronger ability to recover high frequency details and sharp edges.}
\label{vis1}
\end{figure*}
\subsubsection{Instance normalization}
Although batch normalization \cite{ioffe2015batch} has been proved to be effective on many image classification tasks, recent works \cite{DBLP:journals/corr/NahKL16} \cite{ulyanov2016instance} have pointed 
out that batch normalization will decrease the performance of image generation tasks. In \cite{ulyanov2016instance}, the authors proposed to use the instance normalization instead of batch normalization on the image style transform task. Following this, we replace the batch normalization layers 
with instance normalization layers to get better performance on SISR tasks. Instance normalization is to apply the normalization on a single image instead
of the whole batch of images. To introduce the formulation, we denote $x\in R^{T\times C\times W \times H}$ as an input batch which contains $T$ images. Let 
$x_{tkij}$ denote the $tkij-th$ element, where $i$ and $j$ are the spatial dimensions, $k$ is the input feature channel, and $t$ is the index of the image in
the batch. Then the formulation of the instance normalization is given by:
\begin{equation}
y_{tkij} = \dfrac{x_{tkij} - u_{tk}}{\sqrt{\sigma^2_{tk}+\epsilon}}
\end{equation}
where $u_{tk} = \dfrac{1}{HW}\sum_{l = 1}^{W}\sum_{m=1}^{H}x_{tklm}$ and $\sigma_{tk}^2 = \dfrac{1}{HW} \sum_{l = 1}^{W}\sum_{m=1}^{H}(x_{tklm}-u_{tk})^2$.

We replace batch normalization with instance normalization in every layer of the generator $G$. The instance batch normalization layer 
can achieve better performance than batch normalization, and it also can be used for preventing the divergence of the training.
\subsection{Loss Functions}
\label{loss_section}
In this section, we will introduce the formulas of the loss functions.
To get the objective loss functions of discriminator and generator, we need to design adversarial loss, content loss and perceptual loss, respectively. We can get these individual loss functions based on the outputs and intermediate features obtained by discriminator $D$. 
\subsubsection{Adversarial loss}
Our generator $G$ tries to learn a mapping from the image $z$ by bicubic interpolation to the ground truth high resolution image $y$. We design our
discriminator $D$ in a conditional GAN fashion. The adversarial loss function of our GAN model can be expressed as below:
\begin{equation}
\begin{split}
l_a (G,D) = &E_{z,y\sim p_{data}(z,y)}[logD(z,y)]+\\
        &E_{z\sim p_{data}(z)}[log(1-D(z,G(z))]
\end{split}
\end{equation}
In the training phase, the discriminator $D$ tries to minimize this objective function and the generator $G$ tries to maximize it. Compared with the unconditional GAN, the formulation of the adversarial loss function in the unconditional GAN can be expressed as below:
\begin{equation}
\begin{split}
l(G,D) = &E_{y\sim p_{data}(y)}[logD(y)]+\\
        &E_{z\sim p_{data}(z)}[log(1-D(G(z))]
\end{split}
\end{equation}
In contrast, the discriminator of unconditional GAN cannot observe the input bicubic image $z$.

The adversarial loss can encourage our generator to generate the solution that resides on the manifold of the ground truth high resolution images by trying to fool the discriminator. 
\subsubsection{Content Loss}
The adversarial loss can be helpful to recover the high frequency details. Except for the high frequency part, we also need to
design a content loss to ensure the correctness of the low frequency part of the constructed image. The commonly used content loss is the mean square loss. In this paper, we propose to use the Charbonnier loss \cite{DBLP:journals/corr/Barron17} to achieve better performance on the SISR task. 
We denote $y$ as the ground truth high resolution image and $G(z)$ as the constructed image. The Charbonnier loss can be expressed as below:
\begin{equation}
l_y(y,\hat{y}) =E_{z,y\sim p_{data}(z,y)}(\rho(y-G(z))) 
\end{equation}
Where $\rho(x) = \sqrt{x^2+\epsilon^2}$ is the Charbonnier penalty function. To give a comparison, we also try the l1 loss and l2 loss in the experiments.
\begin{figure*}[!h]
\centering
\includegraphics[width = 16cm]{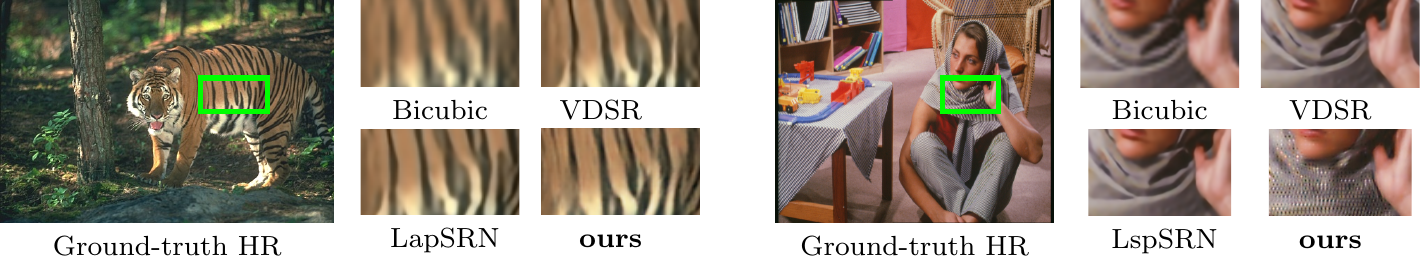}
\caption{Reconstruction results of \textbf{Ours} and recent state-of-the-art methods.[$4\times$ upscaling]}
\label{vis2}
\end{figure*}
\subsubsection{Perceptual Loss}
Previous methods based on the pixel-wise loss always generate images that lack high frequency details. Some perceptual loss based on VGG16 network has been proposed to deal with this issue.
Instead of using the perceptual loss based on the VGG16 classification network, we use the intermediate features of the discriminator to build the perceptual loss. We can get a more robust perceptual loss for image super resolution by that. Additionally, we can reduce the computation budget of the perceptual loss through the reuse of the extracted features obtained by the discriminator. To introduce the
formula of the perceptual loss, we denote $\phi_{i}$ as the feature map computed by the i-th convolution layer(after the activation function layer) within the discriminator. Then, we can define the perceptual loss as:
\begin{equation}
l_p = \sum_{i = 1}^{L}E_{z,y\sim p_{data}(z,y)}(||\phi_{i}(y)-\phi_{i}(G(z))||)
\end{equation}
In this formula, each term in the equation measures the l1 loss of features extracted by i-th layer of the discriminator $D$, where the $G(z)$ represents the image constructed by the generator $G$.
\subsubsection{Optimization}
We use an alternative optimization way to optimize the generator $G$ and discriminator $D$. Based on the individual loss functions presented above, we can define the objective functions
for discriminator $D$ and generator $G$. The formulas are defined as :
\begin{equation}
l_d = -l_a(G,D) + \lambda l_p  
\end{equation}
\begin{equation}
l_g = l_a(G,D) + \lambda_1 l_p + \lambda_2 l_y
\end{equation}
In the training phase, we minimize $l_d$ with respect to the parameters of discriminator $D$ and minimize $l_g$ with respect to the parameters of generator $G$.
To optimize our networks, we alternate between one gradient descent step on $D$ and one step on $G$. For optimizing solver, we use
the ADAM algorithm for both $G$ and $D$. The details can be found in the experimental section.
    \section{Experiments}
    \begin{table*}[h!]
	\centering
	\label{tab:quality} 
	\resizebox{\textwidth}{!}{
		\begin{tabular}{lcccccc}
			\hline
			\multirow{2}{*}{Algorithm}
			&
			\multirow{2}{*}{Scale}
			& 
			\textsc{Set5} & \textsc{Set14} & \textsc{BSDS100} & \textsc{Urban100} & \textsc{manga109} \\
			& 
			&
			PSNR / SSIM  & 
			PSNR / SSIM & 
			PSNR / SSIM & 
			PSNR / SSIM  & 
			PSNR / SSIM  \\
			\hline
			Bicubic & 2 &
			33.65 / 0.930  &
			30.34 / 0.870  &
			29.56 / 0.844  &
			26.88 / 0.841  &
			30.84 / 0.935  \\
			A+~\cite{timofte2014a+} & 2 &
			36.54 / 0.954  &
			32.40 / 0.906  &
			31.22 / 0.887  &
			29.23 / 0.894  &
			35.33 / 0.967  \\
			SRCNN~\cite{Dong:2014td} & 2 &
			36.65 / 0.954  &
			32.29 / 0.903  &
			31.36 / 0.888  &
			29.52 / 0.895  &
			35.72 / 0.968  \\
			FSRCNN~\cite{Dong:2016tw} & 2 &
			36.99 / 0.955 &
			32.73 / 0.909 &
			31.51 / 0.891 &
			29.87 / 0.901 &
			36.62 / 0.971 \\
			SelfExSR~\cite{huang2015single} & 2 &
			36.49 / 0.954 &
			32.44 / 0.906 &
			31.18 / 0.886 &
			29.54 / 0.897 &
			35.78 / 0.968 \\
			RFL~\cite{schulter2015fast} & 2 &
			36.55 / 0.954 &
			32.36 / 0.905 &
			31.16 / 0.885 &
			29.13 / 0.891 &
			35.08 / 0.966 \\
			SCN~\cite{Wang:2015wq} & 2 &
			36.52 / 0.953 &
			32.42 / 0.904 &
			31.24 / 0.884 &
			29.50 / 0.896 &
			35.47 / 0.966 \\
			VDSR~\cite{Kim:2015wv} & 2 &
			37.53 / 0.958 &
			32.97 / \red{0.913} &
			31.90 / \red{0.896} &
			30.77 / \blue{0.914} &
			37.16 / \blue{0.974} \\
			DRCN~\cite{Kim:2016vr} & 2 &
			37.63 / \blue{\underline{0.959}} &
			32.98 / 0.913 &
			31.85 / 0.894 &
			30.76 / 0.913 &
			37.57 / 0.973 \\
			LapSRN\cite{lai2017deep}& 2 &
			37.52 / \red{0.959} &
			33.08 / \red{0.922} &
			31.80 / \blue{0.895} &
			30.41 / 0.910 &
		    37.27 / \red{0.974} \\
            \red{\textbf{SRPGAN (Ours)}} & 2 &
			29.67 / 0.950 &
			27.66 / 0.911 &
			27.89 / \red{0.901} &
			30.41 / 0.892 &
		    37.27 / 0.943 \\
            \hline
			Bicubic & 4 &
			28.42 / 0.810 &
			26.10 / 0.704 &
			25.96 / 0.669 &
			23.15 / 0.659 &
			24.92 / 0.789 \\
			A+~\cite{timofte2014a+} & 4 &
			30.30 / 0.859 &
			27.43 / 0.752 &
			26.82 / 0.710 &
			24.34 / 0.720 &
			27.02 / 0.850 \\
			SRCNN~\cite{Dong:2014td} & 4 &
			30.49 / 0.862 &
			27.61 / 0.754 &
			26.91 / 0.712 &
			24.53 / 0.724 &
			27.66 / 0.858 \\
			FSRCNN~\cite{Dong:2016tw} & 4 &
			30.71 / 0.865 &
			27.70 / 0.756 &
			26.97 / 0.714 &
			24.61 / 0.727 &
			27.89 / 0.859 \\
			SelfExSR~\cite{huang2015single} & 4 &
			30.33 / 0.861 &
			27.54 / 0.756 &
			26.84 / 0.712 &
			24.82 / 0.740 &
			27.82 / 0.865 \\
			RFL~\cite{schulter2015fast} & 4 &
			30.15 / 0.853 &
			27.33 / 0.748 &
			26.75 / 0.707 &
			24.20 / 0.711 &
			26.80 / 0.840 \\
			SCN~\cite{Wang:2015wq} & 4 &
			30.39 / 0.862 &
			27.48 / 0.751 &
			26.87 / 0.710 &
			24.52 / 0.725 &
			27.39 / 0.856 \\
			VDSR~\cite{Kim:2015wv} & 4 &
			31.35 / 0.882 &
			28.03 / 0.770 &
			27.29 / 0.726 &
			25.18 / 0.753 &
			28.82 / 0.886 \\
			DRCN~\cite{Kim:2016vr} & 4 &
			31.53 / \blue{0.884} &
			28.04 / 0.770 &
			27.24 / 0.724 &
			25.14 / 0.752 &
			28.97 / \blue{0.886} \\
			LapSRN\cite{lai2017deep}& 4 &
			31.54 / \red{0.885} &
			28.19 / \blue{0.772}&
			27.32/ \blue{0.728}&
			25.21/ \blue{0.756}&
			29.09 / \red{0.890} \\
            \red{\textbf{SRPGAN(Ours)}} & 4 &
			 22.68 / 0.880 &
			22.50 / \red{0.786} &
			23.91 / \red{0.749} &
			20.00 / \red{0.763} &
			21.00 / 0.860\\
			\hline
			Bicubic & 8 &
			24.39 / 0.657 &
			23.19 / 0.568 &
			23.67 / 0.547 &
			20.74 / 0.515 &
			21.47 / 0.649 \\
			A+~\cite{timofte2014a+} & 8 &
			25.52 / 0.692 &
			23.98 / 0.597 &
			24.20 / 0.568 &
			21.37 / 0.545 &
			22.39 / 0.680 \\
			SRCNN~\cite{Dong:2014td} & 8 &
			25.33 / 0.689 &
			23.85 / 0.593 &
			24.13 / 0.565 &
			21.29 / 0.543 &
			22.37 / 0.682 \\
			FSRCNN~\cite{Dong:2016tw} & 8 &
			25.41 / 0.682 &
			23.93 / 0.592 &
			24.21 / 0.567 &
			21.32 / 0.537 &
			22.39 / 0.672 \\
			SelfExSR~\cite{huang2015single} & 8 &
			25.52 / 0.704 &
			24.02 / 0.603 &
			24.18 / 0.568 &
			21.81 / 0.576 &
			22.99 / 0.718 \\
			RFL~\cite{schulter2015fast} & 8 &
			25.36 / 0.677 &
			23.88 / 0.588 &
			24.13 / 0.562 &
			21.27 / 0.535 &
			22.27 / 0.668 \\
			SCN~\cite{Wang:2015wq} & 8 &
			25.59 / 0.705 &
			24.11 / 0.605 &
			24.30 / 0.573 &
			21.52 / 0.559 &
			22.68 / 0.700 \\
			VDSR~\cite{Kim:2015wv} & 8 &
			25.72 / 0.711 &
			24.21 / 0.609 &
			24.37 / 0.576 &
			21.54 / 0.560 &
			22.83 / 0.707 \\
			LapSRN\cite{lai2017deep}& 8 &
			26.14 / \blue{0.738} &
			24.44 / \blue{0.623} &
			24.54 / \blue{0.586} &
			21.81 / \blue{0.581} &
			23.39 / \red{0.735} \\
            \red{\textbf{SRPGAN(Ours)}} & 8 &
			19.14 / \red{0.743} &
			19.10 / \red{0.635} &
			21.55 / \red{0.613} &
			17.68 / \red{0.607} &
			18.68 / \blue{0.730} \\
			\hline
		\end{tabular}
	}
    \caption{
		Quantitative evaluation of state-of-the-art SR algorithms: average PSNR/SSIM for scale factors $2\times$, $4\times$ and $8\times$.
		\red{\textbf{Red}} text indicates the best SSIM score and \blue{\underline{blue}} text indicates the second best performance of SSIM score.
	}
	\vspace{+0.2cm}
	\vspace{-4mm}
    \label{total_results}
\end{table*}
\subsection{Training Details}
For training dataset, we use images from T91, BSDS200 \cite{MartinFTM01} and General100 datasets. In each training batch, we randomly select 64 image patches as the high resolution patches, with each patch in the size of $128\times 128$. We obtain the low resolution patches by downsampling the high resolution patches using the bicubic kernel with specified downsampling factor. We augment the training data in the following ways: (1) Random Rotation: Randomly rotate the images by $90$ or $180$ degrees. (2) Brightness adjusting:
Randomly adjust the brightness of the images. (3) Saturation adjusting: Randomly adjust the saturation of the images. We pre-process all the images by dividing the image data by $255$. Finally, we get about 640 thousand patches for training. 

We initialize the parameters using "Xavier" \cite{he2016deep}. We train our model from scratch with ADAM optimizer by setting $\beta_1 = 0.9$, $\beta_2 = 0.99$, and $\epsilon = 10^{-8}$. The 
learning rate is initialized as $10^{-4}$ and the learning rate decreased to $10^{-5}$ while we finished $10^6$ iterations.
We set the weight term in the loss function as $\lambda = 0.01$ in equation(6), $\lambda_1 = 1 ~and~\lambda_2 = 1$ in equation(7). 
Our implementation is based on Tensorflow. We have trained 3 models for scaling factor of 2, 4, 8 respectively. It takes about 18 hours for training one model on one GTX1080.
\subsection{Quantitative Evaluation}
We evaluate the performance of our method on five benchmarks: SET5, SET14, BSDS100, URBAN100, and MANGA109. The metrics we used are PSNR and SSIM \cite{wang2004image}. 
We compare the proposed method with previous state-of-the-art SISR methods. For scaling factors, we test our model on 2x,4x and 8x. Table\ref{total_results} shows the overall quantitative comparisons for 2x, 4x and 8x. Most of the results of other methods are cited from \cite{lai2017deep} and \cite{Kim:2015wa}. Because our method dose not optimize the mean square loss directly, the PSNR score of our method is much
lower than other methods. Other than that, our method tends to generate more realistic images and recover more details than the state-of-the-art methods as shown in Figure\ref{vis1}. On the other hand, our method is competitive on the mean SSIM score which has been shown to correlate with human perception on different benchmarks.
Our SRPGAN method performs favorably against existing methods on the most used benchmarks with different scaling factor (2x,4x and 8x).
\vspace{-0.03cm}
From the results, our method has a poor performance on the MANGA dataset, the main reason is that our training dataset consists of real life images and our GAN model tend to reconstruct realistic images, but the MANGA dataset is a dataset consisting of Japanese comics. For other benchmarks, our methods have obvious improvements on SSIM score.

\subsection{Visualizations}
In the context of perceptual quality, our method can recover realistic textures from heavily down-sampling images on the public benchmarks. We have selected some images from the benchmarks to visualization the effective of our method in Figure\ref{vis1} and Figure\ref{vis2}. From the results, the images constructed by our method have shown significant gains in perceptual quality.

We have conducted a series experiments to show the effectiveness of our proposed SISR framework and loss functions. In Figure\ref{vis1}, we compare our method with
previous state-of-the-art methods LapSRN \cite{lai2017deep} and VDSR \cite{Kim:2015wv}. Except for these two methods, we have also compared our method with other CNN based methods. We
just list the results of these two methods due to the page limitation. To better visualize the effectiveness of our method, we selected small regions
which contain rich details in the images to magnify. As the Figure\ref{vis1} shows, our method successfully reconstructs stripes on Zebra's bodies (shown in \textcolor{red}{red} and \textcolor{orange}{orange} boxes). On the other hand, LapSRN and VSDR based on a pixel-wise loss just generate blurry images without stripes in that area. In the leg area(\textcolor{yellow}{yellow} and \textcolor{green}{green} box), the perceptual quality of our method is not good enough, but our method tries to recover the stripe details on the leg, the other methods just construct leg images without any details. We also compare with other CNN based methods, such as SRCNN \cite{Dong:2014td}. In contrast, our approach surpasses other state-of-the-art methods to generate richer texture details. Further examples of perceptual improvements can be found in Figure\ref{vis2}. For these images, we can see that our method has constructed high resolution images with good perceptual quality, those
methods which are based on pixel-wise loss have generated blurry and over-smooth images.
\begin{figure*}[h]
\centering
\includegraphics[width = 14cm]{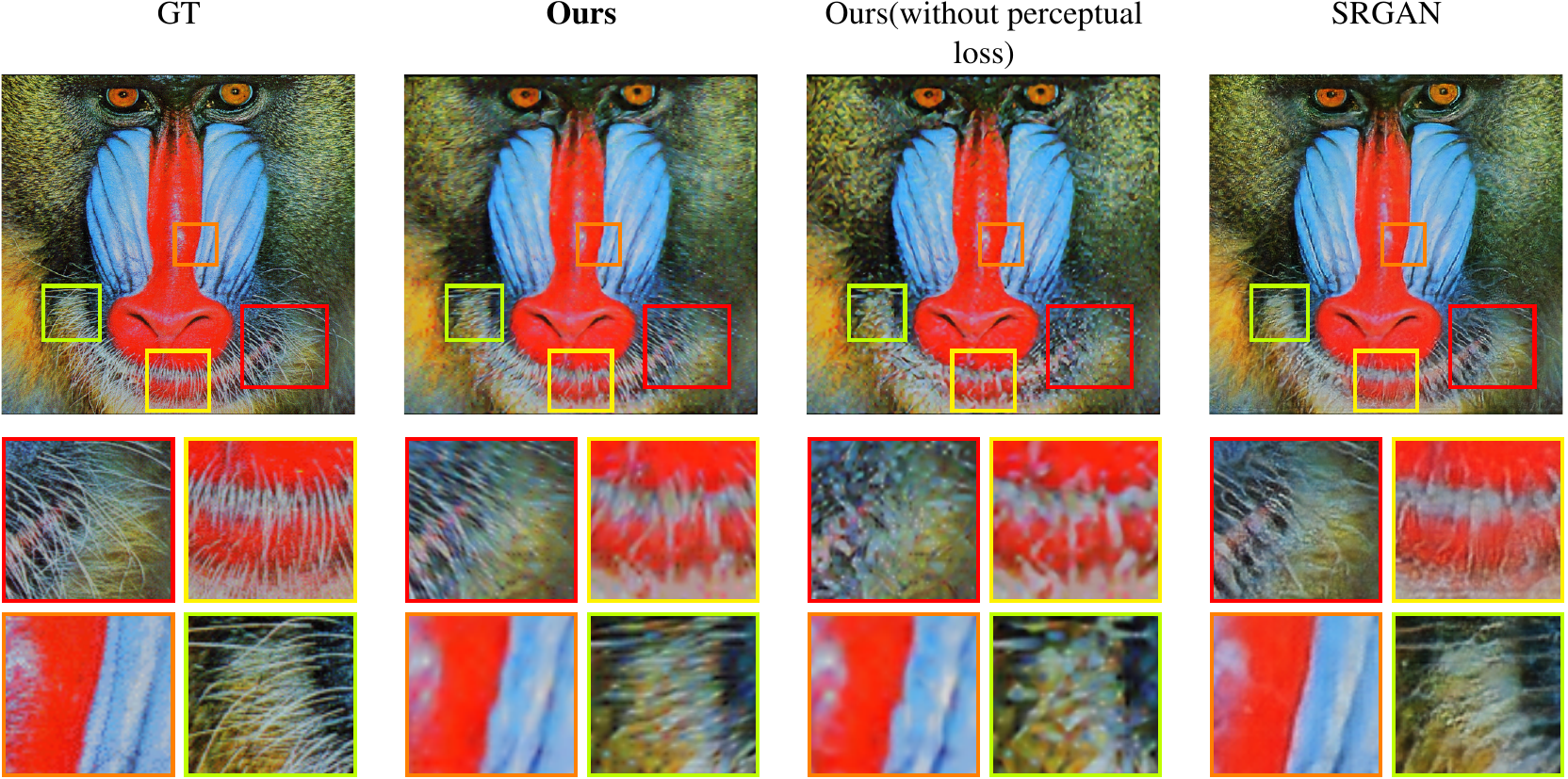}
\caption{Reconstruction results [4x upscaling] of different perceptual loss strategy (our framework with proposed perceptual loss (the second column), our framework without perceptual loss(the third column), SRGAN with VGG perceptual loss(the fourth column))}
\label{vis3}
\end{figure*}
\subsection{Analysis of loss function}
Except for comparison with other methods, we also evaluate the performance of the generator network without our perceptual loss or replacing the Charbonnier loss with
other pixel-wise loss, such as l2 loss. We firstly train a model without perceptual loss. The quantitative results are summarized in Table2 and the visualization results are in Figure\ref{vis3}. From the Table\ref{loss}, the proposed framework with perceptual loss achieves much better performance than trained without perceptual loss. From Figure\ref{vis3} the model trained with perceptual loss has a better perceptual quality than the model trained without perceptual loss. We can observe from the sub images that our method trained with perceptual loss can accurately reconstruct the beards of the baboon(in \textcolor{red}{red} box). 
\begin{table}[H]
\centering
\begin{tabular}{ccc}
Dataset& Perceptual loss&SSIM\\
\hline
Set14& Yes& 0.786\\
Set14& No& 0.754\\
BSDS100& Yes& 0.749\\
BSDS100& No& 0.716\\
\hline
\end{tabular}
\caption{Quantitative comparison of different perceptual loss strategies. We can get a better performance with our robust perceptual loss.}
\label{loss}
\end{table}
We also compare our perceptual loss with the SRGAN which is based on the VGG perceptual loss (see more details in the follow subsection). 

We have conducted further experiments to explore the content loss. To validate the effect of the Charbonnier loss function, we trained a model with l2 content loss respectively. Through the experiments, the model with l2 content loss requires more training epochs to achieve comparable performance than the model trained with the Charbonnier content loss. The results are shown in Table\ref{content_loss}. 
\begin{table}
\centering
\begin{tabular}{cccc}
Dataset& Content Loss&SSIM& Training epochs\\
\hline
Set14& Charbinnier& 0.786& 100\\
Set14& l1& 0.782&100\\
Set14& l2& 0.763&100\\
\hline
\end{tabular}
\caption{Quantitative comparison of different content loss.}
\label{content_loss}
\end{table}
\subsection{Comparison with SRGAN}
We conduct experiments to compare 
our method with the SRGAN\cite{Ledig:2016vc} based on the VGG perceptual loss. We can see that our perceptual loss is more robust than the VGG perceptual loss (see in Figure\ref{vis3}). Note that we have trained a SRGAN model using the open source Tensorflow code on Github\footnote{https://github.com/zsdonghao/SRGAN}. For fair comparison, we train the SRGAN model using the same training dataset as our own method. In the training phase, we train that on a scaling factor of 4x for 100 epochs which is the same as our method.

As we can see in Figure\ref{vis3}, our SRPGAN method does a better job at reconstructing fine details, such as the beards of the baboon ( in \textcolor{red}{red} boxes of the second and the fourth columns), leading to pleasing visual results. On the other hand, in the training phase, our SRPGAN does not need an extra VGG classification net to build the perceptual loss, compared with the SRGAN method. This may help to reduce the computation budget while training.
\subsection{Limitations}
While our model is capable of constructing realistic images with sharp edges and rich details on a large scale factor(4x, 8x). There are still some limitations of our methods. One limitation of our GAN based model is that the constructed images have checker board artifacts at the pixel level. The artifacts are visible in Figure\ref{vis3} upon magnification of the sub image regions. This phenomenon is also mentioned in many previous literatures\cite{odena2016deconvolution}. To solve this issue, one can replace 
the transpose convolution with resize convolution\cite{odena2016deconvolution} and sub-pixel convolution \cite{Shi:2016tm}. We mark this as a part of future work.
    \section{Conclusion}
In this paper, we have highlighted some limitations of the pixel wise loss based methods. To solve these issues, we propose a general framework based on generative and adversarial network (GAN) for single image super resolution. Based on the framework, we design individual loss functions and combined them to form the objective functions for discriminator and generator respectively. Our method achieves the highest SSIM score on most of commonly used benchmarks, and also construct images with better perceptual quality than previous methods, especially for large upscaling factors (4x and 8x). The quantitative and visualization results have shown that SRPGAN surpasses previous methods on details recovering and perceptual quality.
{\small
\bibliographystyle{ieee}
\bibliography{ref}
}

\end{document}